\documentclass{article}
\usepackage{ijcai18}

\usepackage{times}
\usepackage{xcolor}
\usepackage{soul}
\usepackage[utf8]{inputenc}
\usepackage[small]{caption}
\usepackage{epsfig}
\usepackage{graphicx}
\usepackage{amsmath}
\usepackage{amssymb}

\title{View-Volume Network for Semantic Scene Completion from a Single Depth Image}

\author{
Yu-Xiao Guo$^{1,2}$, \,\,\,\,
Xin Tong$^2$ 
\\ 
$^1$ University of Electronic Science and Technology of China \\
$^2$ Microsoft Research Asia \\
yuxiao.guo@outlook.com,
xtong@microsoft.com
}

\begin{document}

\maketitle

\begin{abstract}
We introduce a View-Volume convolutional neural network (VVNet) for inferring the occupancy and semantic labels of a volumetric 3D scene from a single depth image. The VVNet concatenates a 2D view CNN and a 3D volume CNN with a differentiable projection layer. Given a single RGBD image, our method extracts the detailed geometric features from the input depth image with a 2D view CNN and then projects the features into a 3D volume according to the input depth map via a projection layer. After that, we learn the 3D context information of the scene with a 3D volume CNN for computing the result volumetric occupancy and semantic labels. With combined 2D and 3D representations, the VVNet efficiently reduces the computational cost, enables feature extraction from multi-channel high resolution inputs, and thus significantly improves the result accuracy. We validate our method and demonstrate its efficiency and effectiveness on both synthetic SUNCG and real NYU dataset. 

\end{abstract}

\section{Introduction}
Reconstructing and understanding a 3D scene from its partial observations is an important technique in many robotic and vision tasks, such as indoor navigation, object retrieval, and visual reasoning. Given a single depth image captured from a 3D scene, a set of methods recently have been developed  \cite{Song_2016_ssc,Yang_2017_3Dal} for automatically predicting the semantic labels or completing 3D shapes of the objects in the scene using the convolutional neural networks (CNN). To achieve good performance, previous studies have revealed that both local geometric details and global 3D context of the scene need be learned in this task \cite{Song_2016_ssc}. The former one helps the system to identify the small objects in the scene, and the later one is used for inferring occluded objects from the scene layout. However, designing a CNN that can efficiently learn both features is a non-trivial task. 

2D CNN based methods \cite{Gupta2014,guptaCVPR15a} take the depth input as an additional channel of RGB image and apply 2D CNNs for scene segmentation and object detection. Although these methods can fully exploit the high resolution input to generate detailed segmentation results, they ignore the 3D context information of the scene and thus cannot infer the invisible part of the scene. 3D CNN based methods \cite{CVPR15_Wu,thanh2016field,Song_2016_deepslide,Yang_2017_3Dal} convert input depth maps or point clouds into a volumetric representation and design 3D CNNs for 3D scene segmentation or object completion. However, the high computational and memory cost of the 3D CNN limits their capability for recovering object details. Recently, Song et al.\shortcite{Song_2016_ssc} proposed SSCNet for semantic scene completion, where the system simultaneously predicts the object shapes and semantic labels from a single depth image. However, the 3D CNN used in their solution limits the input resolution and the depth of the neural networks, which leads to wrong labels and missing shape details in the results.
  
In this paper, we propose a cascaded convolutional neural network, named View-Volume Net (VVNet), for semantic scene completion from a single depth image. The key idea of our method is to handle the local geometric details and global 3D context with two convolutional neural networks: a 2D view CNN for extracting 2D geometric features, and a 3D volume CNN for learning 3D context of the scene. Given a single depth image captured from a 3D scene, our method first extracts a set of 2D feature maps from the input depth image with the 2D view CNN and then projects the feature maps into a 3D feature volume according to the input depth map via a projection layer. After that, the 3D volume CNN is applied to the 3D feature volume to learn the 3D context information and predicts the occupancy and semantic label of each voxel in the view frustum. Since the feature projection is differentiable, the cascaded VVNet can be trained end-to-end.

The VVNet efficiently learns both geometry features and 3D context from the training dataset. The 2D view CNN avoids the high computational cost and memory consumption of the 3D CNN, which not only enables us to extract geometric features from the high-resolution input depth map, but also allows us to exploit multiple signals computed from the input depth image for feature extraction. Meanwhile, the 3D volume CNN defined over the low resolution feature volume efficiently learns the global 3D context of the scene. Moreover, the VVNet provides a flexible framework for combining variant 2D and 3D CNNs. To this end, we design and evaluate a set of VVNet models with different configurations for semantic scene completion.

We train the VVNets on both synthetic SUNCG dataset and real NYU dataset and validate our design. We also compare their performance with previous methods. Experimental results demonstrate that our method outperforms the state-of-the-art methods on both datasets, with much better accuracy and three times speed up for training, as well as more than 7 times speed up for inference. With different configurations, we further offer VVNet models with different trade-offs between the training/inference cost and result accuracy. All these VVNet models achieve better accuracy than previous solutions.



\section{Related Work}
In this section, we discuss related work and focus on methods for analyzing and completing a 3D scene from depth images or 3D point clouds, as well as the deep learning approaches that are based on hybrid 2D and 3D representations. Please refer to \cite{Ioannidou2017} for a survey of deep learning techniques for 3D data processing. 

\subsection{3D Scene Analysis}
A set of methods have been proposed for scene segmentation, scene completion, and object detection from an input RGBD image or depth image. 2D image-based methods regard the depth as an additional channel of the 2D RGB image and leverage manually-crafted features \cite{guptaCVPR13,atapourdepthcomp2017} or 2D deep neural networks \cite{Gupta2014,guptaCVPR15a} for these scene analysis tasks. 3D volume-based approaches convert the input depth map into a volumetric representation and exploit manually crafted 3D features \cite{Ren_2016_CVPR} or 3D CNNs \cite{Song_2016_deepslide} for detecting 3D objects from the input RGBD image. Although these methods can successfully detect and segment visible 3D objects and scenes in the input RGBD images, they cannot infer the scenes that are totally occluded. Instead, our method predicts semantic labeling and 3D shapes for both visible and invisible objects in a 3D scene.

Liu et al. ~\shortcite{Liu_2017_ICCV} introduced 3DCNN-DQN-RNN for parsing 3D point cloud of a scene. PointNet \cite{QiPointNet16} and PointNet++ \cite{Qi2017PointNet} develop deep learning framework on 3D point cloud for scene semantic labeling and other 3D shape analysis tasks. These methods take the 3D point cloud of whole 3D scene as the input. On the contrary, our method takes a single depth image for semantic scene completion.
 
\subsection{3D Scene Completion}
Firman et al.~\shortcite{FirmanCVPR2016} inferred the occluded 3D object shapes from a single depth image via random forest. Zheng et al.~\shortcite{Zheng13} completed the occluded scene in the input depth image with a set of pre-defined rules and refined the completion results by physical reasoning. These methods perform scene segmentation and completion in two separate steps. Recently, Song et al.~\shortcite{Song_2016_ssc} proposed 3D SSCNet for simultaneously predicting the semantic labels and volumetric occupancy of the 3D objects from a single depth image. Although this method unifies segmentation and completion and significantly improves the result, the expensive 3D CNN limits the input volume resolution and network depth, and thus restrains its performance. By combining 2D CNN and 3D CNN, our method efficiently reduces the training and inference cost, enhances the network depth and thus significantly improves the result accuracy.

\subsection{3D Object Completion}
A set of methods reconstruct the 3D object shape from a single depth image using 3D shape retrieval \cite{Rock_2015_CVPR}, Convolutional Deep Belief Network (CDBN) \cite{CVPR15_Wu,thanh2016field}, or a 3D Generative Adversarial Networks (GAN) \cite{Yang_2017_3Dal}. All these methods model the input depth maps and resulting 3D shapes with a 3D volumetric representation. Although these methods can be combined with other scene segmentation methods for predicting 3D shapes of the visible object in the input depth map, they cannot be used for inferring objects that are totally occluded. Our method is designed for recovering complete 3D shapes of both visible and occluded 3D objects from a single depth image of a 3D scene.

\subsection{Hybrid Representation for 3D Deep Learning}
A set of methods \cite{choy2016,Girdhar16b,Rezende_NIPS2016,drcTulsiani17,Yan2016learning,hspHane17} cascade 2D encoder and 3D decoder for reconstructing 3D object shapes from a single-view color image. Because the feature vector extracted from the 2D color image has no 3D position information, it is mapped to a low resolution 3D volume ($2^3$ or $4^3$) via fully connected (FC) neural networks. In our method, the 2D features extracted from the input depth map inherit the 3D positions of the input depth map and thus can be directly mapped to a 3D volume via projection. Wang et al.~\shortcite{Wang_2017_3Dinpaint} combined 3D GAN and 2D recurrent convolutional networks (RCN) for reconstructing high resolution 3D shapes from corrupted 3D models, where the 2D RCN takes 2D slices of the 3D volume generated by the 3D GAN as input for completing 3D shapes. Our method takes 2D depth map as the input and applies 3D volume CNN to the projected 2D features for 3D scene completion. Kalogerakis et al.~\shortcite{Kalogerakis:2017:ShapePFCN} fused semantic segmentation results generated by 2D fully convolutional networks (FCN) from different views into a conditional random field defined on 3D object surface for object segmentation. Instead of directly obtaining the segmentation results in each view, the 2D View CNN in VVNet only extracts intermediate geometric features for the 3D Volume CNN to learn the 3D context information in the whole volume for scene completion.




\begin{figure*}[t]
	\begin{center}
		\includegraphics[width=\textwidth]{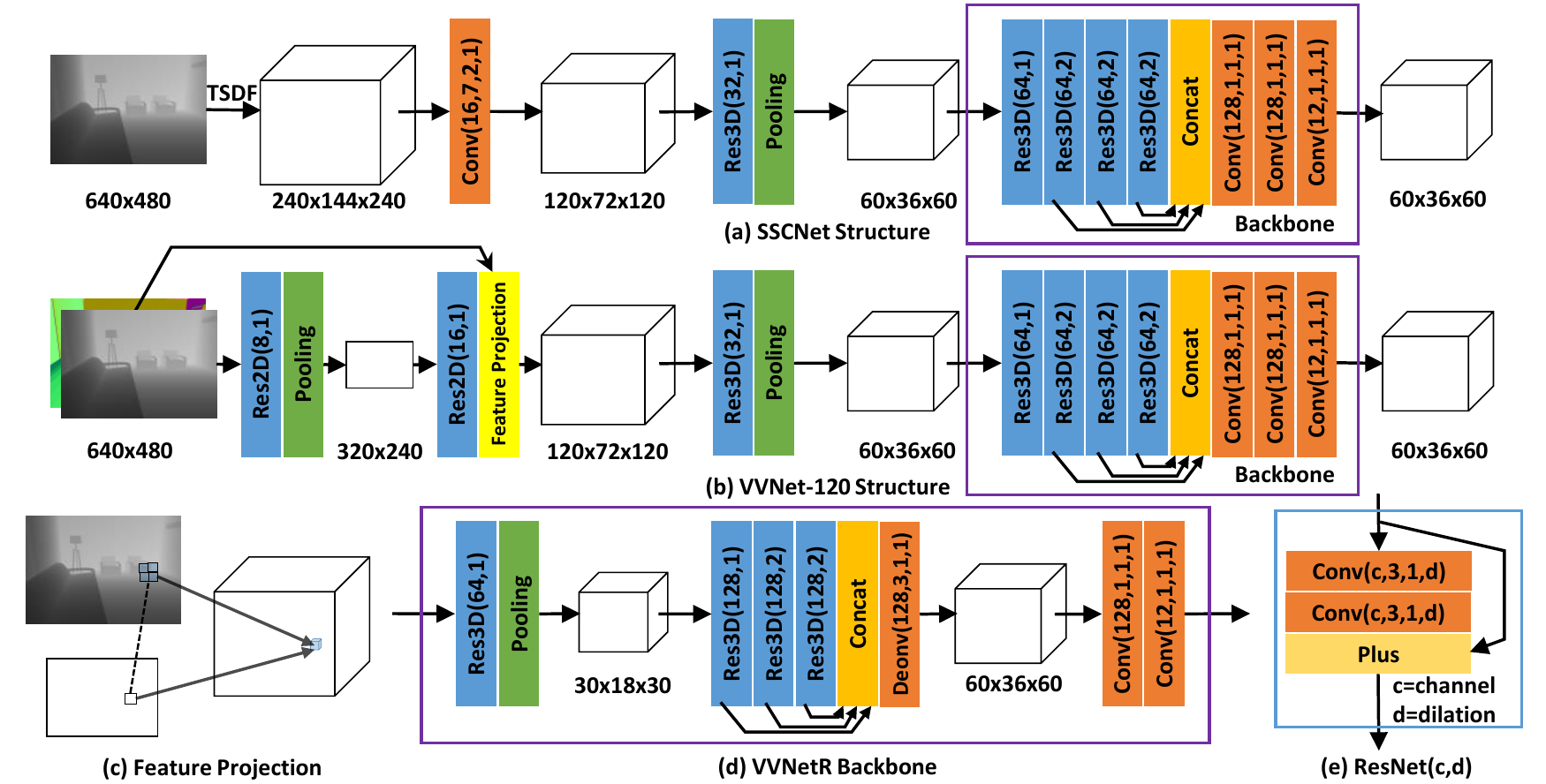}
	\end{center}
	\vspace{-2mm}
	\caption{The network structures of VVNet and SSCNet for semantic scene completion.}
	\vspace{-2mm}
	\label{fig:overview}
\end{figure*}

\section{View-Volume Network}
Given a single-view depth image $I_d$ captured from a 3D scene, our VVNet outputs semantic label $C={c_0, ...c_{N+1}}$ for each voxel of the scene inside the view frustum. Here we follow \cite{Song_2016_ssc} to denote the number of object classes as $N$ and mark all empty voxels by $c_0$. As illustrated in Fig.~\ref{fig:overview}(b), the VVNet consists of three parts: a 2D view network for extracting 2D features from the input depth image, a feature projection layer for converting the 2D features maps into a 3D feature volume, and a 3D volume network for inferring voxel labels from the projected feature volume. In the following part of this section, we discuss the design of each part and VVNet training.

\subsection{2D View Network}
The 2D view network extracts 2D geometry features from the input depth map. For this purpose, our method first computes the normal map from the input depth image and then feeds both normal and depth maps to the 2D view network. As illustrated in Fig.~\ref{fig:overview}(b), we apply the residual neural network (ResNet) structure in our 2D view network design. Each ResNet block includes two convolution layers and a shortcut from input to output as shown in Figure~\ref{fig:overview}(e). When a pooling layer is applied after a ResNet block, the feature map resolution is halved and the number of feature maps is doubled. For 2D ResNet used in the view network, a batch normalization layer is applied after each convolution layer. The total number of the layers in the 2D view network is determined by the resolution of the target feature maps ($320 \times 240$ in Figure~\ref{fig:overview}).

\subsection{Feature Projection}
As shown in Fig.~\ref{fig:overview}(c), the projection layer projects the 2D feature maps constructed by the 2D view network into a 3D feature volume. The voxel size of the feature volume is set to be twice of the average distance of the neighboring depth pixels. Because the feature volume resolution is always lower than the feature map resolution, several neighboring features will be projected into the same voxel. This is equivalent to perform a pooling operation during the projection.

To project the 2D feature maps into the feature volume, we first construct an axis-aligned volume in the viewing coordinate system as in \cite{Song_2016_ssc} and then upsample the feature maps to input depth resolution with the nearest neighboring sampling. After that, we project the feature vectors of all depth pixels into the voxels of the 3D feature volume according to their viewing directions and depth values. Finally, we get the feature vector in each voxel by averaging the feature vectors projected into it. For the voxels that are not occupied by any depth image pixels, their feature vectors are zero. We found that other pooling operations (e.g. max pooling) can also be used for computing the feature vector in each voxel but have the similar affect to the result. Instead of downsampling the depth image to the feature map resolution, we apply this super-sampling scheme for feature projection so that we can avoid the holes caused by the perspective projection and large variations in the depth image. During training, we record the mapping between feature map pixels and voxels in a table for gradient back-propagation.

In our current implementation, we set the resolutions of our feature volumes to be same as the ones in the SSCNet \cite{Song_2016_ssc} for a fair comparison. We set the resolution of the feature map that corresponds to the largest 3D volume ($240 \times 144 \times 240 $) to be $640 \times 480$ because for the TSDF in this resoluton, the one constructed from half-resolution depth image ($320 \times 240$) is almost same as the one constructed from $640 \times 480$ depth image. For the downsampled feature volumes, we scale down the resolutions of the feature maps accordingly.

\subsection{3D Volume Network}
After the feature projection layer, the view-dependent 2D feature maps extracted by the view CNN are converted to a view-independent 3D feature volume. In this step, we extract the 3D context of the scene and infer the semantic label of voxels from the 3D feature volume via a 3D volume CNN. To this end, we follow the SSCNet in our 3D feature volume CNN design(Figure~\ref{fig:overview}(b)), where the 3D features are first extracted by the ResNet layers and then fed into a backbone network to generate the semantic labels for all voxels. We also design a new backbone network with enlarged reception field. As shown in Figure~\ref{fig:overview}(d), the new backbone adds a new pooling layer to downsample the extracted 3D features and then applies the original backbone network to 3D features at a lower resolution. After that, we deconvolute the concatenated features to the output resolution and generate volumetric semantic labels using two convolution layers. Note that in 3D volume network, we do not apply batch normalization after each convolution layer. We denote the VVNet with the SSCNet backbone as \emph{VVNet}, and the one with the new backbone as \emph{VVNetR}. In Section~\ref{sec:evaluation}, we demonstrate that our VVNetR not only reduces the computational cost for learning the 3D context, but also improves the result accuracy.

\begin{figure}[t]
	\begin{center}
		\includegraphics[]{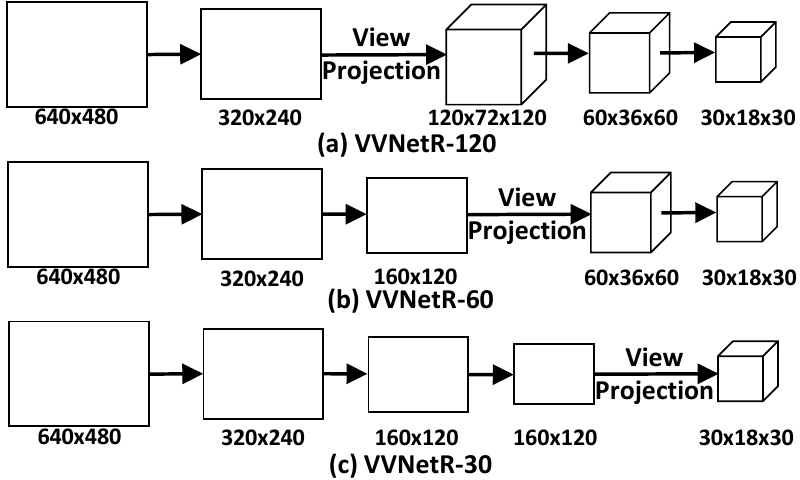}
	\end{center}
	\vspace{-2mm}
	\caption{Different trade-offs between view and volume networks in VVNetR design.}
	\vspace{-2mm}
	\label{fig:VVNetTradeoff}
\end{figure}

\subsection{Trade-off between View and Volume Networks}
The VVNet provides a flexible and general framework for combining 2D and 3D CNN for 3D scene analysis. By choosing different resolutions of the result feature maps of 2D view network, we can make different trade-offs between the depth of 2D view network and 3D volume network. Fig.~\ref{fig:VVNetTradeoff} illustrates three VVNets, which are named as VVNetR-120, VVNetR-60, and VVNetR-30, where the numbers in the name indicate the resolution of the projected feature volume. On one side, as the resolution of the projected feature volume decreases, more layers in the 3D volume CNN are replaced by the corresponding 2D view network layers, which results in less computations and smaller memory footprints in both network training and inference. On the other side, as more 3D volume network layers are replaced by the 2D view network layers, more detailed 3D context information in the scene may be lost in the projected 3D feature volume and thus leads to degradation of the result accuracy. We evaluate this trade-off in details in the next section.

\subsection{Network Training}
Given the training data set (i.e the depth images and ground truth volumetric object labels of 3D scene), the VVNet can be trained end-to-end. For this purpose, we use the voxel-wise softmax as the loss function as in \cite{Song_2016_ssc} in the network training. To compute the loss function, we remove all empty voxels in the visible free space, outside field of view and outside the room but include all non-empty voxels and occluded empty voxels. We do not apply the data balancing scheme in \cite{Song_2016_ssc} in our training process.


\section{Evaluation}\label{sec:evaluation}
\begin{table*}
    \begin{center}
    \begin{tabular}{p{2.6cm}|p{.45cm}p{.45cm}p{.45cm}|p{.45cm}p{.45cm}p{.45cm}p{.45cm}p{.45cm}p{.45cm}p{.45cm}p{.45cm}p{.45cm}p{.45cm}p{.45cm}p{.45cm}}
    \hline
    & \multicolumn{3}{c|}{scene completion} & \multicolumn{12}{c}{semantic scene completion}\\
    \hline
    Network                &prec. &recall& IoU  &ceil. &floor & wall & win. &chair & bed  & sofa &table & tvs  &furn. &objs. & avg. \\
    \hline
    SSCNet                 & 76.3 & \textbf{95.2} & 73.5 & 96.3 & 84.9 & 56.8 & 28.2 & 21.3 & 56.0 & 52.7 & 33.7 & 10.9 & 44.3 & 25.4 & 46.4 \\
    SSCNet$^*$             & 90.4 & 89.7 & 82.0 & 97.8 & \textbf{88.2} & 59.4 & 37.3 & 39.2 & 77.9 & 68.9 & 48.3 & 31.5 & 56.8 & 44.9 & 59.1 \\
    SSCNet$^*$-half        & 90.5 & 89.5 & 81.9 & 97.8 & 88.0 & 60.8 & 34.8 & 39.8 & 77.5 & 69.5 & 47.8 & 29.8 & 56.0 & 44.8 & 58.8 \\
    \hline
    VVNet-120-half         & 90.7 & 89.6 & 82.1 & 97.9 & 85.2 & 59.4 & 47.5 & 44.2 & 77.4 & 71.1 & 49.3 & 34.2 & 58.2 & 49.0 & 61.3 \\
    VVNet-120-depth        & 90.6 & 89.6 & 82.0 & 97.6 & 84.8 & 58.6 & 44.5 & 44.8 & 77.6 & 70.7 & 48.8 & 33.2 & 57.8 & 46.2 & 60.4 \\
    VVNet-120              & \textbf{90.8} & 90.0 & 82.5 & 97.9 & 85.4 & 58.6 & 49.2 & 45.3 & 79.2 & 71.8 & 50.3 & 37.3 & 62.0 & 50.9 & 62.5 \\
    VVNetR-120            & \textbf{90.8} & 91.7 & 84.0 & \textbf{98.4} & 87.0 & \textbf{61.0} & \textbf{54.8} & \textbf{49.3} & \textbf{83.0} & \textbf{75.5} & \textbf{55.1} & \textbf{43.5} & \textbf{68.8} & \textbf{57.7} & \textbf{66.7} \\
    VVNetR-60             & 90.6 & 92.5 & \textbf{83.7} & 97.6 & 86.7 & 60.2 & 54.4 & 47.2 & 80.7 & 75.0 & 53.8 & 39.4 & 66.9 & 56.1 & 65.3 \\
    VVNetR-30             & 88.8 & 90.2 & 81.0 & 98.0 & 86.4 & 55.6 & \textbf{54.8} & 41.8 & 78.0 & 72.1 & 48.7 & 31.6 & 63.2 & 51.8 & 62.0 \\
    \hline
    \end{tabular}
    \end{center}
    \vspace{-2mm}
    \caption{Performances of different variant VVNet design on the SUNCG dataset. \textbf{half} refers to the network that takes half-resolution image as input. \textbf{depth} refers to the network that use depth only as input.}
    \vspace{-2mm}
    \label{Comparison-SUNCG}
\end{table*}

\begin{table*}
    \begin{center}
    \begin{tabular}{p{3.6cm}|p{.45cm}p{.45cm}p{.45cm}|p{.45cm}p{.45cm}p{.45cm}p{.45cm}p{.45cm}p{.45cm}p{.45cm}p{.45cm}p{.45cm}p{.45cm}p{.45cm}p{.45cm}}
    \hline
    & \multicolumn{3}{c|}{scene completion} & \multicolumn{12}{c}{semantic scene completion}\\
    \hline
    Method                 &prec. &recall& IoU  &ceil. &floor & wall & win. &chair & bed  & sofa &table & tvs  &furn. &objs. & avg. \\
    \hline
    \cite{lin2013holistic} & 58.5 & 49.9 & 36.4 &  0.0 & 11.7 & 13.3 & 14.1 & 9.4  & 29.0 & 24.0 & 6.0  & 7.0  & 16.2 &  1.1 & 12.0\\
    \cite{geiger2015joint} & 65.7 & 58.0 & 44.4 & 10.2 & 62.5 & 19.1 & 5.8  & 8.5  & 40.6 & 27.7 & 7.0  & 6.0  & 22.6 &  5.9 & 19.6\\
    \hline
    SSCNet                 & 59.3 & \textbf{92.9} & 56.6 & 15.1 & 94.6 & 24.7 & 10.8 & 17.3 & 53.2 & 45.9 & 15.9 & 13.9 & 31.1 & 12.6 & 30.5\\
    SSCNet$^*$             & 69.7 & 81.3 & 59.8 & 16.1 & \textbf{94.8} & 27.0 & 10.1 & \textbf{20.6} & 53.2 & 50.1 & 16.7 & \textbf{14.3} & 35.5 & 13.0 & 31.9 \\
    VVNet-120              & 68.4 & 83.2 & 60.0 & 19.2 & 94.4 & 27.2 & \textbf{13.8} & 19.1 & 54.0 & 49.3 & 17.1 & 11.2 & 35.3 & 12.4 & 32.1\\
    VVNetR-120             & \textbf{69.8} & 83.1 & \textbf{61.1} & 19.3 & \textbf{94.8} & 28.0 & 12.2 & 19.6 & \textbf{57.0} & 50.5 & \textbf{17.6} & 11.9 & \textbf{35.6} & 15.3 & 32.9\\
    VVNetR-60              & 68.3 & 85.1 & 60.9 & \textbf{21.6} & 94.5 & \textbf{28.6} & 12.9 & 19.7 & 56.3 & \textbf{51.0} & 17.2 & 10.4 & 35.2 & \textbf{15.6} & \textbf{33.0}\\
    \hline 
    \end{tabular}
    \end{center}
    \vspace{-2mm}
    \caption{The performances of different scene completion methods on the NYU dataset.}
    \vspace{-2mm}
    \label{Comparison-NYU}
\end{table*}

We have implemented VVNet in TensorFlow under Ubuntu 16.04 on a workstation with an Intel 6700K CPU and two NVIDIA GTX 1080Ti GPUs. We use SGD optimization in VVNet training, where we set the momentum as 0.9, learning rate as 0.01, and weight decay as 0.0005. We found that the original SSCNet is not fully trained. For a fair comparison, we ported SSCNet in TensorFlow and trained it with more iterations (150K iteration with batch size 4). We denote our SSCNet implementation as SSCNet$^*$ in the following discussions.

\vspace{-1mm}
\paragraph{Datasets} We validate our method on the synthetic SUNCG \cite{Song_2016_ssc} dataset and the real NYU \cite{Silberman2012} dataset. The SUNCG dataset consists of about 45K synthetic scenes. We select the same training/test dataset used in \cite{Song_2016_ssc} for our network training and evaluation. Specifically, the training dataset includes nearly 150K depth images and corresponding ground truth volume, sampled from a 8K subset of the scenes. The test dataset is sampled from 170 scenes and consists of totally 470 pairs of depth image and ground truth volume. For SUNCG, we train VVNet with 150K iterations and change the learning rate to 0.001 after 100K iterations. We evaluate the results every 2000 steps after 130K iterations, and average them as the final results.

The real NYU dataset includes 1449 depth images (795 for training, 654 for test), captured by the Kinect depth sensor. The ground truth completion and segmentation notations are from \cite{GuoZH15}. Because some manually labeled volumes and their corresponding depth images are not well aligned in the NYU dataset, we also use NYUCAD dataset in \cite{FirmanCVPR2016} in our experiments, in which the depth map is rendered from the label volume. For both NYU and NYUCAD datasets, we fine tune the VVNet models trained from SUNCG dataset with 4K iterations. After that, we test the models at every 200 iterations and pick the best one as the final result.

\vspace{-1mm}
\paragraph{Error Metric} For each neural network, we measure the precision, recall, and IOU of all test results as in \cite{Song_2016_ssc}. The IOU measures the overlapped ratio between intersection and union of the positive prediction volume and the ground truth volume. For scene completion (SC) task , the ground truth volume includes all the occluded voxels in the view frustum. For semantic scene completion (SSC) task, the ground truth volume includes both occluded voxels and visible surface voxels.

\subsection{Ablation Test}
We validate our VVNet design with a set of ablation tests on the SUNCG dataset.

\vspace{-1mm}
\paragraph{Does Higher Image Resolution Help?} We downsample the depth input images to half resolution ($320 \times 240$) and use them for training a SSCNet$^*$-half model and a VVNet-120-half model. As shown in Table.~\ref{Comparison-SUNCG}, the limited TSDF resolution in SSCNet cannot preserve the geometric details in high resolution input image and thus leads to very similar results for both SSCNet$^*$ and SSCNet$^*$-half. Note that the TSDF resolution in SSCNet cannot be increased anymore due to the large memory and computational cost of 3D CNN. On the contrary, our method can fully exploit the high resolution input. Compared to  VVNet-120-half, the IOUs of the VVNet-120 for SC and SSC tasks improve $0.4\%$ and $1.2\%$ respectively.

\vspace{-1mm}
\paragraph{Does Multi-channel Input Help?} To validate the contribution of the normal map to the VVNet result, we train VVNet-120-depth with the depth image only as input. Compared to VVNet-120 that takes both depth and normal as the input, the IOUs of the VVNet-120-depth decreases $0.5\%$ and $2.1\%$ for SC and SSC tasks respectively, which demonstrates that the normal input helps VVNet to learn the local geometric features. Note that these extra non-depth features (e.g. normal, RGB, etc.) are difficult to be used in 3D CNN training and inference as discussed in several previous methods \cite{Song_2016_ssc,dai2017scancomplete,guedes2018semantic}.

\vspace{-1mm}
\paragraph{Does Larger Reception Field Help?} Table.~\ref{Comparison-SUNCG} compares the performances of VVNet-120 and VVNetR-120. For SC and SSC tasks, our new backbone with larger reception field in VVNetR-120 provides $1.9\%$ and $5.4\%$ IOU improvements compared to VVNet-120.

\begin{table}
    \begin{center}
    \begin{tabular}{l|c|c|c}
    \hline
    & \multicolumn{2}{c|}{training} & \multicolumn{1}{c}{inference}\\
    \hline
    Network           &  memory & speed & speed \\
    \hline
    SSCNet$^*$        & 852M  & 912ms & 578ms \\
    \hline
    VVNet-120         & 846M  & 386ms & 75ms  \\
    VVNetR-120,      & 712M  & 375ms & 74ms  \\
    VVNetR-60,       & 336M  & 194ms & 51ms  \\
    VVNetR-30,       & 246M  & 156ms & 45ms  \\
    \hline    
    \end{tabular}    
    \end{center}
    \vspace{-2mm}
    \caption{Memory footprints and computational times of different networks for model training and inference.}
    \vspace{-2mm}
    \label{Comparison-Performance}
\end{table}

\vspace{-1mm}
\paragraph{Trade-offs between View and Volume Networks} We compare the performances of VVNet models with different combinations of 2D view and 3D volume CNNs. As shown in Table.~\ref{Comparison-SUNCG}, the IOUs of the VVNetR-60 is slightly worse than the IOUs of the VVNetR-120 ($0.3\%$ and $1.4\%$ decreasing for SC and SSC respectively). However, the VVNetR-60 only requires half of memory footprint and $70\%$ computational time that the VVNetR-120 needs in the training. For VVNetR-30, we observe a relatively large performance drop from the VVNetR-120. A possible reason is that the low resolution feature volume projected from the 2D features lose too much detailed 3D context information.

\subsection{Evaluation}
we test the performance of VVNet network for semantic scene completion task on all three datasets and compare our method with other existing approaches.

\begin{figure*}
    \begin{center}
    \includegraphics[width=.9\textwidth]{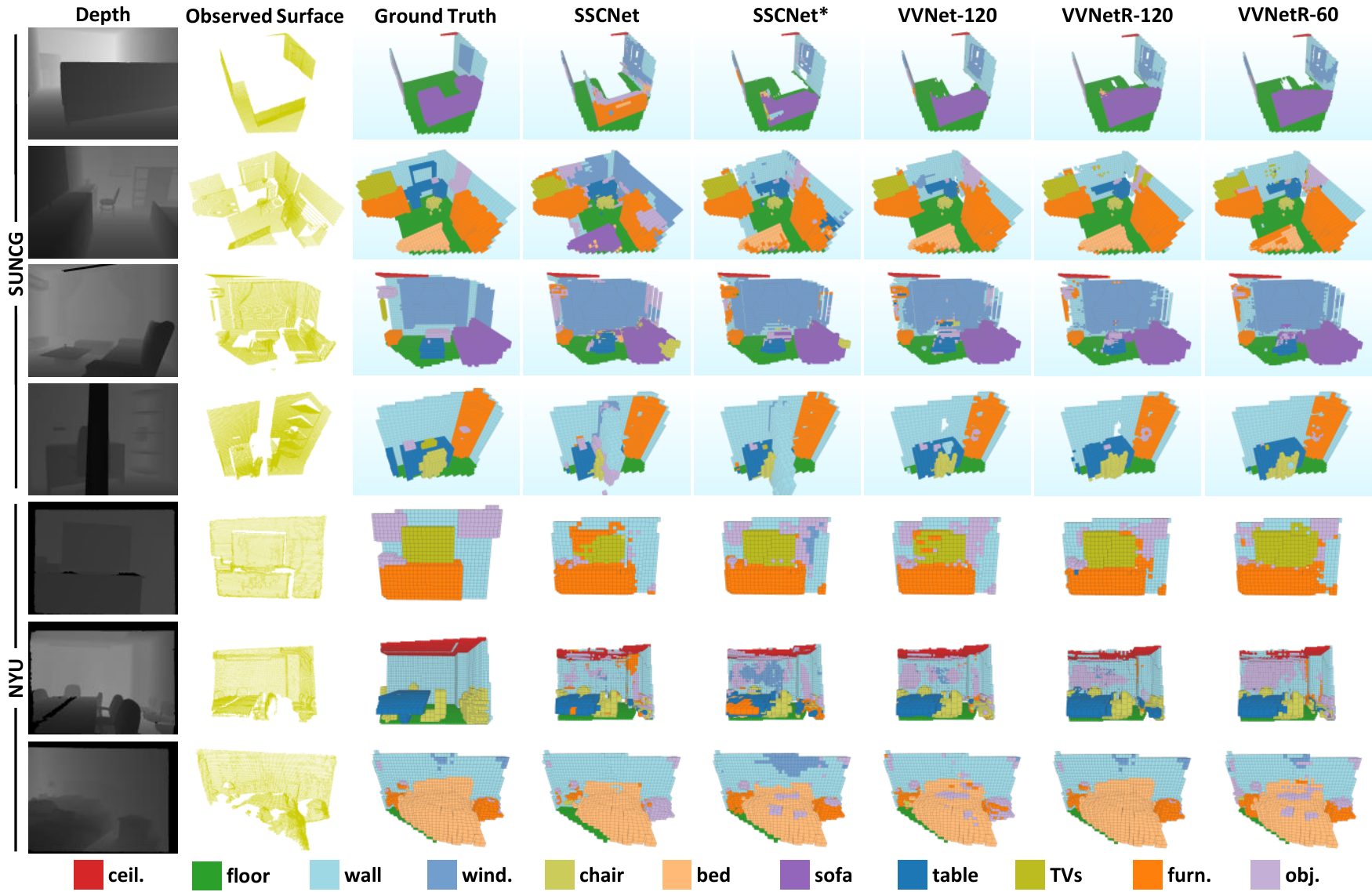}
    \end{center}
    \caption{Semantic scene completion results generated by different methods for SUNCG and NYU datasets.}
    \vspace{-2mm}
    \label{Visual-SUNCG-Comparison}
    \vspace{-2mm}
\end{figure*}

\begin{table}
    \begin{center}
    \begin{tabular}{l|ccc}
    \hline
    Method                 & prec. & recall & IoU \\
    \hline
    \cite{Zheng13}         & 60.1  & 46.7   & 34.6\\
    \cite{FirmanCVPR2016}  & 66.5  & 69.7   & 50.8\\
    \hline
    SSCNet                 & 75.0  & 96.0   & 73.0\\
    SSCNet$^*$             & 83.2  & 92.7   & 78.0\\
    VVNet-120              & 83.3  & 93.1   & 78.5\\
    VVNetR-120            & 86.4  & 92.0   & 80.3\\
    VVNetR-60             & 85.6  & 91.5   & 79.2\\
    \hline
    \end{tabular}
    \end{center}
    \vspace{-2mm}
    \caption{Performances of different methods on NYUCAD dataset.}
    \vspace{-2mm}
    \label{Comparison-NYUCAD}
\end{table}

\vspace{-1mm}
\paragraph{SUNCG} For SUNCG dataset, we compare the IOUs of VVNetR-120, SSCNet, and SSCNet$^*$ for both SC and SSC tasks. As shown in Table.~\ref{Comparison-SUNCG}, the VVNetR-120 achieves the best performance in both SC and SSC tasks. Compared to SSCNet$^*$, the IOUs of VVNetR-120 increase $1.4\%$ and $5.9\%$ for SC and SSC tasks respectively. Table.~\ref{Comparison-SUNCG} also lists the IOU for each object class. Note that for all object class except floor, the VVNetR-120 achieves the best results. Fig.~\ref{Visual-SUNCG-Comparison} illustrates the semantic scene completion results generated by different methods.


\vspace{-1mm}
\paragraph{NYU \& NYUCAD} We compare our method with SSCNet$^*$ and other existing methods on both NYU and NYUCAD dataset. Table ~\ref{Comparison-NYU} and ~\ref{Comparison-NYUCAD} list the performances of these methods on both datasets. For both NYU and NYUCAD datasets, our VVNet achieves the best performance among all the methods \cite{lin2013holistic,geiger2015joint}. We believe that the small performance gap between our VVNetR models and SSCNet$^*$ on the NYU dataset is caused by the misalignment between the input and output.

\vspace{-1mm}
\paragraph{Memory Footprint \& Computational Cost} Compared to the SSCNet that is based on 3D CNN, our VVNet combines the 2D CNN and 3D CNN and thus significantly reduces both memory footprint and computational cost in training and inference. Table.\ref{Comparison-Performance} compares the memory footprints and computational costs of different models for training and inference. Compared to SSCNet, the VVNetR-120 provides much better accuracy and three-times speed up for training, as well as more than 7 times speed up for inference. With $40\%$ memory footprint of SSCNet in training, the VVNetR-60 offers 4.7 times speed up for training and more than 10 times speed up for inference at the cost of slightly degraded accuracy compared to VVNetR-120. Note that the accuracy of VVNetR-60 is still better than the SSCNet.

\section{Conclusion}
We introduce a view-volume CNN for semantic scene completion from a single depth image. Our method concatenates a 2D view CNN and a 3D volume CNN with a projection layer and can be trained end-to-end. The VVNet provides a general and flexible framework for fusing 2D and 3D CNNs for efficient 3D learning. We validate our method on both synthetic and real datasets. Results shown that our method significantly improves the result accuracy, reduces the computational cost in training and inference, and offers variant trade-offs between the cost and accuracy.

For the future work, it is interesting to explore the trade-offs between 2D CNNs and 3D CNNs and find solutions to improve the performance of VVNet with less 3D volume layers. Another interesting direction is to extend our view-volume framework for multiple view CNN solutions.


\bibliographystyle{named}
\small
\bibliography{egbib}

\begin{thebibliography}{}

\bibitem[\protect\citeauthoryear{Atapour-Abarghouei and
  Breckon}{2017}]{atapourdepthcomp2017}
A.~Atapour-Abarghouei and T.P. Breckon.
\newblock Depthcomp: Real-time depth image completion based on prior semantic
  scene segmentation.
\newblock In {\em BMVC}, pages 1--13, 2017.

\bibitem[\protect\citeauthoryear{Choy \bgroup \em et al.\egroup
  }{2016}]{choy2016}
Christopher~B Choy, Danfei Xu, JunYoung Gwak, Kevin Chen, and Silvio Savarese.
\newblock 3d-r2n2: A unified approach for single and multi-view 3d object
  reconstruction.
\newblock In {\em ECCV}, pages 628--644, 2016.

\bibitem[\protect\citeauthoryear{Dai \bgroup \em et al.\egroup
  }{2017}]{dai2017scancomplete}
Angela Dai, Daniel Ritchie, Martin Bokeloh, Scott Reed, J{\"u}rgen Sturm, and
  Matthias Nie{\ss}ner.
\newblock Scancomplete: Large-scale scene completion and semantic segmentation
  for 3d scans.
\newblock {\em arXiv preprint arXiv:1712.10215}, 2017.

\bibitem[\protect\citeauthoryear{Firman \bgroup \em et al.\egroup
  }{2016}]{FirmanCVPR2016}
Michael Firman, Oisin Mac~Aodha, Simon Julier, and Gabriel~J. Brostow.
\newblock Structured prediction of unobserved voxels from a single depth image.
\newblock In {\em CVPR}, pages 5431--5440, 2016.

\bibitem[\protect\citeauthoryear{Geiger and Wang}{2015}]{geiger2015joint}
Andreas Geiger and Chaohui Wang.
\newblock Joint 3d object and layout inference from a single rgb-d image.
\newblock In {\em GCPR}, pages 183--195, 2015.

\bibitem[\protect\citeauthoryear{Girdhar \bgroup \em et al.\egroup
  }{2016}]{Girdhar16b}
R.~Girdhar, D.F. Fouhey, M.~Rodriguez, and A.~Gupta.
\newblock Learning a predictable and generative vector representation for
  objects.
\newblock In {\em ECCV}, pages 484--499, 2016.

\bibitem[\protect\citeauthoryear{Guedes \bgroup \em et al.\egroup
  }{2018}]{guedes2018semantic}
Andre Bernardes~Soares Guedes, Teofilo~Emidio de~Campos, and Adrian Hilton.
\newblock Semantic scene completion combining colour and depth: preliminary
  experiments.
\newblock {\em arXiv preprint arXiv:1802.04735}, 2018.

\bibitem[\protect\citeauthoryear{Guo \bgroup \em et al.\egroup
  }{2015}]{GuoZH15}
Ruiqi Guo, Chuhang Zou, and Derek Hoiem.
\newblock Predicting complete 3d models of indoor scenes.
\newblock {\em arXiv preprint arXiv:1504.02437}, 2015.

\bibitem[\protect\citeauthoryear{Gupta \bgroup \em et al.\egroup
  }{2013}]{guptaCVPR13}
Saurabh Gupta, Pablo Arbelaez, and Jitendra Malik.
\newblock Perceptual organization and recognition of indoor scenes from {RGB-D}
  images.
\newblock In {\em CVPR}. 2013.

\bibitem[\protect\citeauthoryear{Gupta \bgroup \em et al.\egroup
  }{2014}]{Gupta2014}
Saurabh Gupta, Ross Girshick, Pablo Arbel{\'a}ez, and Jitendra Malik.
\newblock Learning rich features from rgb-d images for object detection and
  segmentation.
\newblock In {\em ECCV}, pages 345--360, 2014.

\bibitem[\protect\citeauthoryear{Gupta \bgroup \em et al.\egroup
  }{2015}]{guptaCVPR15a}
Saurabh Gupta, Pablo~Andr{\'{e}}s Arbel{\'{a}}ez, Ross~B. Girshick, and
  Jitendra Malik.
\newblock Aligning {3D} models to {RGB-D} images of cluttered scenes.
\newblock In {\em CVPR}, pages 4731--4740, 2015.

\bibitem[\protect\citeauthoryear{H{\"a}ne \bgroup \em et al.\egroup
  }{2017}]{hspHane17}
Christian H{\"a}ne, Shubham Tulsiani, and Jitendra Malik.
\newblock Hierarchical surface prediction for 3d object reconstruction.
\newblock In {\em arXiv preprint arXiv:1704.00710}. 2017.

\bibitem[\protect\citeauthoryear{Ioannidou \bgroup \em et al.\egroup
  }{2017}]{Ioannidou2017}
Anastasia Ioannidou, Elisavet Chatzilari, Spiros Nikolopoulos, and Ioannis
  Kompatsiaris.
\newblock Deep learning advances in computer vision with 3d data: A survey.
\newblock {\em ACM Computing Surveys}, 50(2):20:1--20:38, 2017.

\bibitem[\protect\citeauthoryear{Jimenez~Rezende \bgroup \em et al.\egroup
  }{2016}]{Rezende_NIPS2016}
Danilo Jimenez~Rezende, S.~M.~Ali Eslami, Shakir Mohamed, Peter Battaglia, Max
  Jaderberg, and Nicolas Heess.
\newblock Unsupervised learning of 3d structure from images.
\newblock In {\em NIPS}, pages 4996--5004. 2016.

\bibitem[\protect\citeauthoryear{Kalogerakis \bgroup \em et al.\egroup
  }{2017}]{Kalogerakis:2017:ShapePFCN}
Evangelos Kalogerakis, Melinos Averkiou, Subhransu Maji, and Siddhartha
  Chaudhuri.
\newblock 3{D} shape segmentation with projective convolutional networks.
\newblock In {\em CVPR}, pages 3779--3788, 2017.

\bibitem[\protect\citeauthoryear{Lin \bgroup \em et al.\egroup
  }{2013}]{lin2013holistic}
Dahua Lin, Sanja Fidler, and Raquel Urtasun.
\newblock Holistic scene understanding for 3d object detection with rgbd
  cameras.
\newblock In {\em ICCV}, pages 1417--1424, 2013.

\bibitem[\protect\citeauthoryear{Liu \bgroup \em et al.\egroup
  }{2017}]{Liu_2017_ICCV}
Fangyu Liu, Shuaipeng Li, Liqiang Zhang, Chenghu Zhou, Rongtian Ye, Yuebin
  Wang, and Jiwen Lu.
\newblock 3dcnn-dqn-rnn: A deep reinforcement learning framework for semantic
  parsing of large-scale 3d point clouds.
\newblock In {\em ICCV}, pages 5678--5687, 2017.

\bibitem[\protect\citeauthoryear{Nguyen \bgroup \em et al.\egroup
  }{2016}]{thanh2016field}
D.~T. Nguyen, B.~S. Hua, M.~K. Tran, Q.~H. Pham, and S.~K. Yeung.
\newblock A field model for repairing 3d shapes.
\newblock In {\em CVPR}, pages 5676--5684, 2016.

\bibitem[\protect\citeauthoryear{Qi \bgroup \em et al.\egroup
  }{2016}]{QiPointNet16}
Charles~Ruizhongtai Qi, Hao Su, Kaichun Mo, and Leonidas~J. Guibas.
\newblock Pointnet: Deep learning on point sets for 3d classification and
  segmentation.
\newblock In {\em CVPR}, pages 652--660, 2016.

\bibitem[\protect\citeauthoryear{Qi \bgroup \em et al.\egroup
  }{2017}]{Qi2017PointNet}
Charles~Ruizhongtai Qi, Li~Yi, Hao Su, and Leonidas~J. Guibas.
\newblock Pointnet++: Deep hierarchical feature learning on point sets in a
  metric space.
\newblock In {\em NIPS}, pages 5105--5114, 2017.

\bibitem[\protect\citeauthoryear{Ren and Sudderth}{2016}]{Ren_2016_CVPR}
Zhile Ren and Erik~B. Sudderth.
\newblock Three-dimensional object detection and layout prediction using clouds
  of oriented gradients.
\newblock In {\em CVPR}, pages 1525--1533, 2016.

\bibitem[\protect\citeauthoryear{Rock \bgroup \em et al.\egroup
  }{2015}]{Rock_2015_CVPR}
Jason Rock, Tanmay Gupta, Justin Thorsen, JunYoung Gwak, Daeyun Shin, and Derek
  Hoiem.
\newblock Completing 3d object shape from one depth image.
\newblock In {\em CVPR}, pages 2484--2493, 2015.

\bibitem[\protect\citeauthoryear{Silberman \bgroup \em et al.\egroup
  }{2012}]{Silberman2012}
Nathan Silberman, Derek Hoiem, Pushmeet Kohli, and Rob Fergus.
\newblock Indoor segmentation and support inference from rgbd images.
\newblock In {\em ECCV}, pages 746--760, 2012.

\bibitem[\protect\citeauthoryear{Song and Xiao}{2016}]{Song_2016_deepslide}
Shuran Song and Jianxiong Xiao.
\newblock {D}eep {S}liding {S}hapes for amodal 3{D} object detection in {RGB-D}
  images.
\newblock In {\em CVPR}, pages 808--816, 2016.

\bibitem[\protect\citeauthoryear{Song \bgroup \em et al.\egroup
  }{2017}]{Song_2016_ssc}
Shuran Song, Fisher Yu, Andy Zeng, Angel~X Chang, Manolis Savva, and Thomas
  Funkhouser.
\newblock Semantic scene completion from a single depth image.
\newblock In {\em CVPR}, pages 1746--1754, 2017.

\bibitem[\protect\citeauthoryear{Tulsiani \bgroup \em et al.\egroup
  }{2017}]{drcTulsiani17}
Shubham Tulsiani, Tinghui Zhou, Alexei~A. Efros, and Jitendra Malik.
\newblock Multi-view supervision for single-view reconstruction via
  differentiable ray consistency.
\newblock In {\em CVPR}, pages 2626--2634, 2017.

\bibitem[\protect\citeauthoryear{Wang \bgroup \em et al.\egroup
  }{2017}]{Wang_2017_3Dinpaint}
Weiyue Wang, Qiangui Huang, Suya You, Chao Yang, and Ulrich Neumann.
\newblock Shape inpainting using 3d generative adversarial network and
  recurrent convolutional networks.
\newblock In {\em ICCV}, pages 2298--2306, 2017.

\bibitem[\protect\citeauthoryear{Wu \bgroup \em et al.\egroup
  }{2015}]{CVPR15_Wu}
Zhirong Wu, Shuran Song, Aditya Khosla, Linguang Zhang, Xiaoou Tang, and
  Jianxiong Xiao.
\newblock 3d shapenets: A deep representation for volumetric shape modeling.
\newblock In {\em CVPR}, pages 1912--1920, 2015.

\bibitem[\protect\citeauthoryear{Yan \bgroup \em et al.\egroup
  }{2016}]{Yan2016learning}
X.~Yan, J.~Yang, E.~Yumer, Y.~Guo, and H.~Lee.
\newblock Perspective transformer nets: Learning single-view 3d object
  reconstruction without 3d supervision.
\newblock In {\em NIPS 2016}, pages 1696--1704, 2016.

\bibitem[\protect\citeauthoryear{Yang \bgroup \em et al.\egroup
  }{2017}]{Yang_2017_3Dal}
Bo~Yang, Hongkai Wen, Sen Wang, Ronald Clark, Andrew Markham, and Niki Trignoi.
\newblock 3d object reconstruction from a single depth view with adversarial
  learning.
\newblock In {\em ICCV}, pages 679--688, 2017.

\bibitem[\protect\citeauthoryear{Zheng \bgroup \em et al.\egroup
  }{2013}]{Zheng13}
Bo~Zheng, Yibiao Zhao, Joey~C. Yu, Katsushi Ikeuchi, and Song-Chun Zhu.
\newblock Beyond point clouds: Scene understanding by reasoning geometry and
  physics.
\newblock In {\em CVPR}, pages 3127--3134, 2013.

\end{thebibliography}

\end{document}